\documentclass{article}

\PassOptionsToPackage{numbers, compress}{natbib}
\usepackage[final]{nips_2017}

\usepackage{hyperref}       
\usepackage{url}            
\usepackage{booktabs}       
\usepackage{amsfonts}       
\usepackage{amssymb}        
\usepackage{nicefrac}       
\usepackage{microtype}      

\usepackage[textsize=tiny]{todonotes}

\usepackage[ruled,vlined,linesnumbered]{algorithm2e}
\usepackage{amsmath}
\usepackage{amssymb}
\usepackage{bm}
\usepackage{graphicx}
\graphicspath{{figs/}}
\usepackage{mathtools}
\usepackage{subcaption}
\newcommand\ci{\perp\!\!\!\perp}
\DeclareMathOperator*{\argmin}{argmin}
\DeclareMathOperator*{\argmax}{argmax}

\hypersetup{
    pdftitle={Beyond normality: Learning sparse probabilistic graphical models in the non-Gaussian
    setting},
    pdfauthor={Rebecca E.~Morrison, Ricardo Baptista, Youssef Marzouk},
    pdfkeywords={sparsity, non-Gaussian, graphical models, Markov structure, transport maps,
    structure learning, sparse graphs, NIPS}
}

\title{Beyond normality: Learning sparse probabilistic graphical models in the non-Gaussian setting}

\author{
  Rebecca E.~Morrison
    \\
  MIT\\
  \texttt{rmorriso@mit.edu} \\
  \And
  Ricardo Baptista\\
  MIT\\
  \texttt{rsb@mit.edu} \\
  \And
  Youssef Marzouk\\
  MIT\\
  \texttt{ymarz@mit.edu} \\
}

\begin{document}

\maketitle

\begin{abstract}
We present an algorithm to identify sparse dependence structure in continuous and
non-Gaussian probability distributions, given a corresponding set of data. The conditional
independence structure of an arbitrary distribution can be represented as an undirected graph (or
Markov random field), but most algorithms for learning this structure are restricted to the discrete or
Gaussian cases. Our new approach allows for more realistic and accurate descriptions of the
distribution in question, and in turn better estimates of its sparse Markov structure. Sparsity in the
graph is of interest as it can accelerate inference, improve sampling methods, and reveal important
dependencies between variables. The algorithm relies on exploiting the connection between the
sparsity of the graph and the sparsity of transport maps, which deterministically couple one
probability measure to another.
\end{abstract}

\section{Undirected probabilistic graphical models}
Given $n$ samples from the joint probability distribution of some
random variables $X_1, \ldots, X_p$,
%
%
%
%
a common goal is to discover the underlying Markov random field. This field
is specified by an undirected graph $G$, comprising the set of
vertices $V=\{1, \ldots, p\}$ and the set of edges $E$. The edge set
$E$ encodes the conditional independence structure of the distribution, i.e., $e_{jk}
\notin E \iff X_j \ci X_k \, | \, \bm{X}_{V \setminus \{jk\}}$. Finding the edges of the graph is
useful for several reasons: knowledge of conditional independence relations can accelerate
inference and improve sampling methods, as well as illuminate structure underlying the process that
generated the data samples. This problem---identifying an undirected graph given samples---is quite
well studied for Gaussian or discrete distributions. In the Gaussian setting, the inverse covariance, or
precision, matrix precisely encodes the sparsity of the graph. That is, a zero appears in the
$jk$-th entry of the precision if and only if variables $X_j$ and $X_k$ are
conditionally independent given the rest. Estimation of the support of the precision matrix is often
achieved using a maximum likelihood estimate with $\ell_1$ penalties. Coordinate descent (glasso)
\citep{friedman2008sparse} and neighborhood selection \citep{meinshausen2006high} algorithms can be
consistent for sparse recovery with few samples, i.e., $p > n$. In the discrete setting,
\cite{loh2012structure} showed that for some particular graph structure, the support of a
generalized covariance matrix encodes the conditional independence structure of the graph,
while \citep{ravikumar2010high} employed sparse $\ell_1$-penalized logistic regression to
identify Ising Markov random fields.

Many physical processes, however, generate data that are continuous but non-Gaussian. One example is
satellite images of cloud cover formation, which may greatly impact weather conditions and climate
\citep{sengupta2016predictive,perron2013climatology}. Other examples
include biological processes such as bacteria growth \citep{ghosh2016anomalous},
heartbeat behavior \citep{peng1993long}, and transport in biological tissues
\citep{liu2004characterizing}. 
Normality assumptions about
the data may prevent the detection of important conditional dependencies. Some approaches have
allowed for non-Gaussianity, such as the nonparanormal approach of
\citep{liu2009nonparanormal,liu2012high}, which uses copula
functions to estimate a joint non-Gaussian density while preserving conditional independence.
However, this approach is still restricted by the choice of copula function, and as far as we know,
no fully general approach is yet available. Our goal in this work is to {\it consistently}
estimate graph structure when the underlying data-generating process is non-Gaussian. To do so, we
propose the algorithm \textsc{SING} (Sparsity Identification in Non-Gaussian distributions).
\textsc{SING} uses the framework of transport maps to characterize arbitrary continuous
distributions, as described in \S\ref{sec:tm}. Our representations of transport maps employ
polynomial expansions of degree $\beta$. Setting $\beta = 1$ (i.e., linear maps) is equivalent to
assuming that the data are well approximated by a multivariate Gaussian. With $\beta > 1$,
\textsc{SING} acts as a generalization of Gaussian-based algorithms by allowing for arbitrarily rich
parameterizations of the underlying data-generating distribution, without additional assumptions on
its structure or class.

\section{Learning conditional independence}\label{sec:ci}
Let 
$X_1, \ldots, X_p$ have a smooth and
strictly positive density $\pi$ on $\mathbb{R}^p$. A
recent result in \citep{spantini2017inference} shows that 
conditional independence of the random variables $X_j$ and $X_k$ can be determined
as follows:
\begin{equation} X_j \ci X_k \, \vert \, \bm{X}_{V \setminus \{jk\}} \iff \partial_{jk} \log
    \pi(\bm{x}) = 0,\,\, \forall \,\bm{x} \in {\mathbb{R}^p},
\end{equation}
where $\partial_{jk}(\cdot)$ denotes the $jk$-th mixed partial derivative.
Here, we define the {\it generalized precision} as the matrix $\Omega$, where
$\Omega_{jk} = \mathbb{E}_\pi \left[ \left| \partial_{jk} \log \pi(\bm{x})
\right|\right]$. Note that $\Omega_{jk} = 0$ is a sufficient condition that variables
$X_j$ and $X_k$ be conditionally independent. Thus, finding the zeros in the matrix
$\Omega$ is equivalent to finding the graphical structure underlying the density $\pi$. Note that
the zeros of the precision matrix in a Gaussian setting encode the same information---the lack of an
edge---as the zeros in the generalized precision introduced here.

Our approach identifies the zeros of $\Omega$ and thus the edge set
$E$ in an iterative fashion via the algorithm \textsc{SING},
outlined in
\S\ref{sec:sing}. Note that computation of an entry of the generalized
precision relies on an expression for the density $\pi$. We
represent $\pi$ and also estimate $\Omega$ using the notion of a
\textit{transport map} between probability distributions.  This map
is estimated from independent samples $\bm{x}^{(i)} \sim \pi, i=1,\dots,n$, as described in
\S\ref{sec:tm}. Using a map, of the particular form described below,
offers several advantages: (1) computing the generalized precision
given the map is efficient (e.g., the result of a convex optimization
problem); (2) the transport map itself enjoys a notion of sparsity
that directly relates to the Markov structure of the data-generating
distribution; (3) a coarse map may capture these Markov properties
without perfectly estimating the high-dimensional density $\pi$.

Let us first summarize our objective and proposed approach. We aim to solve the
following graph recovery problem:
\begin{quote}\vspace{-.1cm}
Given samples $\{\bm{x}^{(i)} \}_{i=1}^{n}$ from some unknown distribution,
   find 
   the dependence graph of this distribution and the associated Markov properties.
\end{quote}
Our proposed approach loosely follows these steps:
    \begin{itemize}\vspace{-.1cm}\setlength\itemsep{-.1em}
        \item Estimate a transport map from samples
        \item Given an estimate of the map, compute the generalized precision $\Omega$
        \item Threshold $\Omega$ to identify a (sparse) graph
        \item Given a candidate graphical structure, re-estimate the map. Iterate\dots
\end{itemize}

The final step---re-estimating the map, given a candidate graphical structure---makes use of a
strong connection between the sparsity of the graph and the sparsity of the transport map (as shown
by \citep{spantini2017inference} and described in \S\ref{ssec:spar-tm}). Sparsity in the
graph allows for sparsity in the map, and a sparser map allows for improved estimates of $\Omega$.
This is the motivation behind an iterative algorithm.

\section{Transport maps}\label{sec:tm}

The first step of \textsc{SING} is to estimate a transport map from samples \citep{marzouk2016sampling}. A
transport map induces a deterministic coupling of two probability distributions
\citep{rosenblatt1952remarks,moselhy2012bayesian,parno2016multiscale,spantini2017inference}.
Here, we build a map between the distribution generating the samples (i.e., $\bm{X} \sim
\pi$) and a standard normal distribution $\eta = \mathcal{N}(0,I_p)$. 
As described in
\citep{villani2008optimal, bogachev2005triangular}, given two distributions with smooth and strictly positive densities ($\pi$,
$\eta$),\footnote{Regularity assumptions on $\pi$, $\eta$ can be substantially relaxed, though
\eqref{eq:push} and \eqref{eq:pull} may need modification \cite{bogachev2005triangular}.} there
exists a monotone map $S: \mathbb{R}^p \to \mathbb{R}^p$ such that $S_\sharp \pi = \eta$ and
$S^\sharp\eta = \pi$, where
\begin{align}
    S_\sharp \pi(\bm{y}) &= \pi \circ S^{-1}(\bm{y})\det\left(\nabla S^{-1}(\bm{y})\right) \label{eq:push}\\
    S^\sharp \eta(\bm{x}) &= \eta \circ S(\bm{x})\det\left(\nabla S(\bm{x})\right). \label{eq:pull}
\end{align}
    We say $\eta$ is the {\it pushforward} density of $\pi$ by the map $S$, and similarly, $\pi$ is
    the {\it pullback} of $\eta$ by $S$.
    Many possible transport maps satisfy the measure
    transformation conditions above.
 In this work, we restrict our attention to lower triangular monotone increasing maps.
    \citep{rosenblatt1952remarks,knothe1957contributions,bogachev2005triangular} show that, under the conditions above,
    there exists a unique lower triangular map $S$ of the form
    \[ S(\bm{x}) =
\begin{bmatrix*}[l]
    S^1(x_1)\\S^2(x_1,x_2)\\S^3(x_1,x_2,x_3)\\\,\,\vdots\\S^p(x_1,\dots
\dots,x_p)\end{bmatrix*} ,\]
with $\partial_k S^k > 0$. The qualifier ``lower triangular'' refers to the property that each component of the map $S^k$ only depends on variables $x_1,\dots,x_k$. The space of such maps is denoted $\mathcal{S}_\Delta$.

As an example, consider a normal random variable: $\bm{X} \sim \mathcal{N}(0,\Sigma)$. Taking the Cholesky decomposition of the covariance $K K^T =
\Sigma$, then $K^{-1}$ is a linear operator that maps (in distribution) $\bm{X}$ to a random variable $\bm{Y} \sim
\mathcal{N}(0,I_p)$, and similarly, $K$ maps
$\bm{Y}$ to $\bm{X}$. In this example, we associate the map $K^{-1}$ with $S$, since it maps the more exotic distribution
to the standard normal:
\[S(\bm{X}) = K^{-1}\bm{X} \stackrel{d}{=} \bm{Y}, \quad\quad S^{-1}(\bm{Y}) = K\bm{Y} \stackrel{d}{=} \bm{X}.\]
In general, however, the map $S$ may be nonlinear. This is exactly what
allows us to represent and capture arbitrary non-Gaussianity in the samples. The
monotonicity of each component of the map (that is, $\partial_k S^k > 0$) can be enforced
by using the following parameterization:
\[ S^k(x_1,\dots,x_k) = c_k(x_1,\dots,x_{k-1}) + \int_0^{x_k}
\exp{\left\{h_k \left(x_1,\dots,x_{k-1},t\right)\right\}}dt,\]
with functions $c_k: \mathbb{R}^{k-1} \to \mathbb{R}$ and $h_k: \mathbb{R}^{k} \to
\mathbb{R}$. Next, a particular form for $c_k$ and $h_k$ is specified; in this work, we use a linear
expansion with Hermite polynomials for $c_k$ and Hermite functions for $h_k$. An important choice is
the maximum degree $\beta$ of the polynomials. With higher degree, the computational difficulty of the
algorithm increases by requiring the estimation of more coefficients in the expansion. This
trade-off between higher degree (which captures more possible nonlinearity) and
computational expense is a topic of current research \citep{bigoni2017computation}. The space of
lower triangular maps, parameterized in this way, is denoted
$\mathcal{S}_\Delta^\beta$. Computations in the transport map
framework are performed using the software
TransportMaps \citep{tmv1}.

\subsection{Optimization of map coefficients is an MLE problem}
Let $\bm{\alpha} \in \mathbb{R}^{n_\alpha}$ be the vector of coefficients that parameterize the functions
$c_k$ and $h_k$, which in turn define a particular instantiation of the transport map
$S_{\bm{\alpha}}\in \mathcal{S}_\Delta^\beta$. (We include the subscript in this subsection to emphasize that the map depends on
its particular parameterization, but later drop it for notational efficiency.)
To complete the estimation of $S_{\bm{\alpha}}$, it remains to optimize for the coefficients $\bm{\alpha}$. This optimization is achieved by minimizing
the Kullback-Leibler divergence between the density in question, $\pi$, and the pullback of the
standard normal $\eta$ by the map $S_{\bm{\alpha}}$ \cite{tmv1}:
\begin{align}
    \bm{\alpha}^* &= \argmin_{\bm{\alpha}}\mathcal{D}_{KL}\left(\pi || S_{\bm{\alpha}}^\sharp \eta\right)\\
             &= \argmin_{\bm{\alpha}}\mathbb{E}_\pi\left(\log \pi - \log S_{\bm{\alpha}}^\sharp \eta\right)\\
             &\approx \argmax_{\bm{\alpha}}\frac{1}{n}\sum_{i=1}^n \log \left(S_{\bm{\alpha}}^\sharp
             \eta\left(\bm{x}^{(i)}\right)\right) = \hat{\bm{\alpha}}. \label{eq:mle}
\end{align}
As shown in \citep{marzouk2016sampling,parno2014transport}, for standard Gaussian $\eta$ and lower
triangular $S$, this optimization problem is convex and separable across
dimensions $1, \ldots, p$. Moreover, by line~(\ref{eq:mle}), the solution to the optimization
problem is a maximum likelihood estimate $\hat{\bm{\alpha}}$. Given that the $n$ samples are
random, $\hat{\bm{\alpha}}$ converges in distribution as $n \rightarrow \infty$ to a normal random
variable whose mean is the exact minimizer $\bm{\alpha}^*$, and whose variance is
$I^{-1}(\bm{\alpha}^*)/n$, where $I(\alpha)$ is the Fisher information matrix. That is:
\begin{equation} \hat{\bm{\alpha}} \sim
\mathcal{N}\left(\bm{\alpha}^*, \frac{1}{n} I^{-1}(\bm{\alpha}^*)\right), \text{ as } n \rightarrow
\infty.\label{eq:mle-norm}\end{equation}

Optimizing for the map coefficients yields a representation of the density $\pi$ as $S_{\bm{\alpha}}^\sharp
\eta$. Thus, it is now possible to compute the conditional independence scores with the generalized precision:
\begin{align} \Omega_{jk} = \mathbb{E}_\pi\left[\left |\partial_{jk} \log
    \pi(\bm{x})\right|\right]
                          &= \mathbb{E}_\pi\left[\left|\partial_{jk} \log S_{\bm{\alpha}}^\sharp \eta(\bm{x})
                          \right|\right]                     \\
                          &\approx \frac{1}{n}\sum_{i=1}^n\left|\partial_{jk} \log S_{\bm{\alpha}}^\sharp
                          \eta\left(\bm{x}^{(i)}\right)
                          \right| = \hat{\Omega}_{jk}.\label{eq:ci}
\end{align}

The next step is to threshold $\hat{\Omega}$. First, however, we
explain the connection between the two notions of sparsity---one of the graph and the other of the
map.

\subsection{Sparsity and ordering of the transport map}\label{ssec:spar-tm} 
\begin{figure}
  \centering
    \begin{subfigure}{.45\textwidth}
  \includegraphics[width=.8\textwidth]{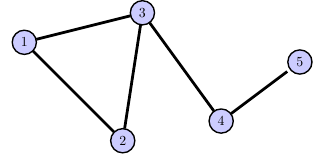}
        \caption{\label{fig:spar-tm-a}}
    \end{subfigure}
    \begin{subfigure}{.45\textwidth}
  \includegraphics[width=.8\textwidth]{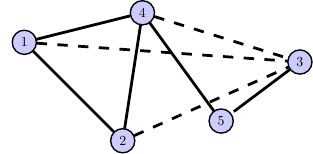}
        \caption{\label{fig:spar-tm-b}}
    \end{subfigure}
    \caption{(a) A sparse graph with an optimal ordering; (b) Suboptimal ordering induces extra
    edges.\label{fig:spar-tm}}
\end{figure}

Because the transport
maps are lower triangular, they are in some sense already sparse. However, it may be possible to
prescribe more sparsity in the form of the map. \citep{spantini2017inference} showed that the Markov
structure associated with the density $\pi$ yields tight lower bounds on the sparsity pattern $\mathcal{I}_S$,
where the latter is defined as the set of all pairs $(j,k), j<k$, such that the $k$th component of
the map does not depend on the $j$th variable: $\mathcal{I}_S \coloneqq \{(j,k):j<k, \partial_j S^k
= 0 \}.$ The variables associated with the complement of this set are called {\it active}. Moreover,
these sparsity bounds can be identified by simple graph operations; see \S5 in
\citep{spantini2017inference} for details. Essentially these operations amount to identifying the
intermediate graphs produced by the variable elimination algorithm, but they do \emph{not} involve
actually performing variable elimination or marginalization. The process starts with node $p$,
creates a clique between all its neighbors, and then ``removes'' it. The process continues in the
same way with nodes $p-1$, $p-2$, and so on until node $1$. The edges in the resulting (induced)
graph determine the sparsity pattern of the map $\mathcal{I}_S$. In general, the induced graph will
be more highly connected unless the original graph is chordal. Since the set of added edges, or
fill-in, depends on the ordering of the nodes, it is beneficial to identify an ordering that
minimizes it. For example, consider the graph in Figure~\ref{fig:spar-tm-a}.
The corresponding map has a nontrivial sparsity
pattern, and is thus more sparse than a dense lower triangular map:
\begin{equation} S(\bm{x}) =
\begin{bmatrix*}[l]
    S^1(x_1)\\S^2(x_1,x_2)\\S^3(x_1,x_2,x_3)\\S^4(\quad\quad\;\,\,\,x_3,x_4)\\S^5(\quad\quad
\quad\quad\,x_4,x_5)\end{bmatrix*} ,\quad\quad \mathcal{I}_S =
\{(1,4),(2,4),(1,5),(2,5),(3,5)\}.\end{equation}
Now consider Figure~\ref{fig:spar-tm-b}. Because of the suboptimal ordering, edges must be added to the
induced graph, shown in dashed lines. The resulting map is then less sparse than in
\ref{fig:spar-tm-a}: $\mathcal{I}_S = \{(1,5),(2,5)\}$.

An ordering of the variables is equivalent to a permutation $\varphi$, but the problem of finding an
optimal permutation is NP-hard, and so we turn to heuristics.  Possible schemes include so-called
min-degree and min-fill \citep{koller2009probabilistic}.  Another that we have found to be
successful in practice is reverse Cholesky, i.e., the reverse of a good ordering for sparse Cholesky
factorization \citep{saad2003iterative}. We use this in the examples below.

\begin{figure}[!hb]
  \centering
    \begin{subfigure}{.45\textwidth}
        \centering
  \includegraphics[width=.8\textwidth]{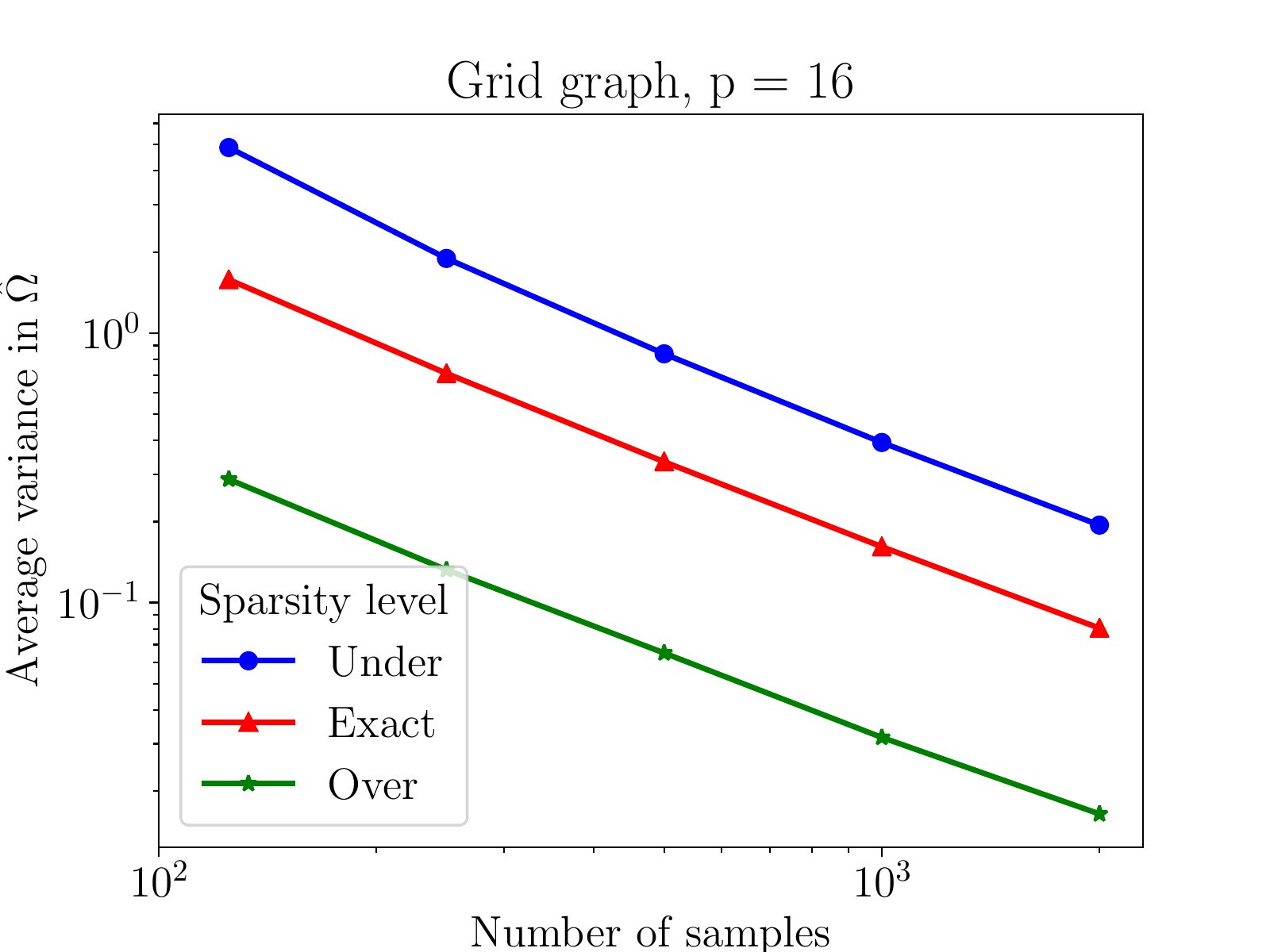}
        \caption{}
    \end{subfigure}\hspace{1em}
    \begin{subfigure}{.45\textwidth}
        \centering
  \includegraphics[width=.8\textwidth]{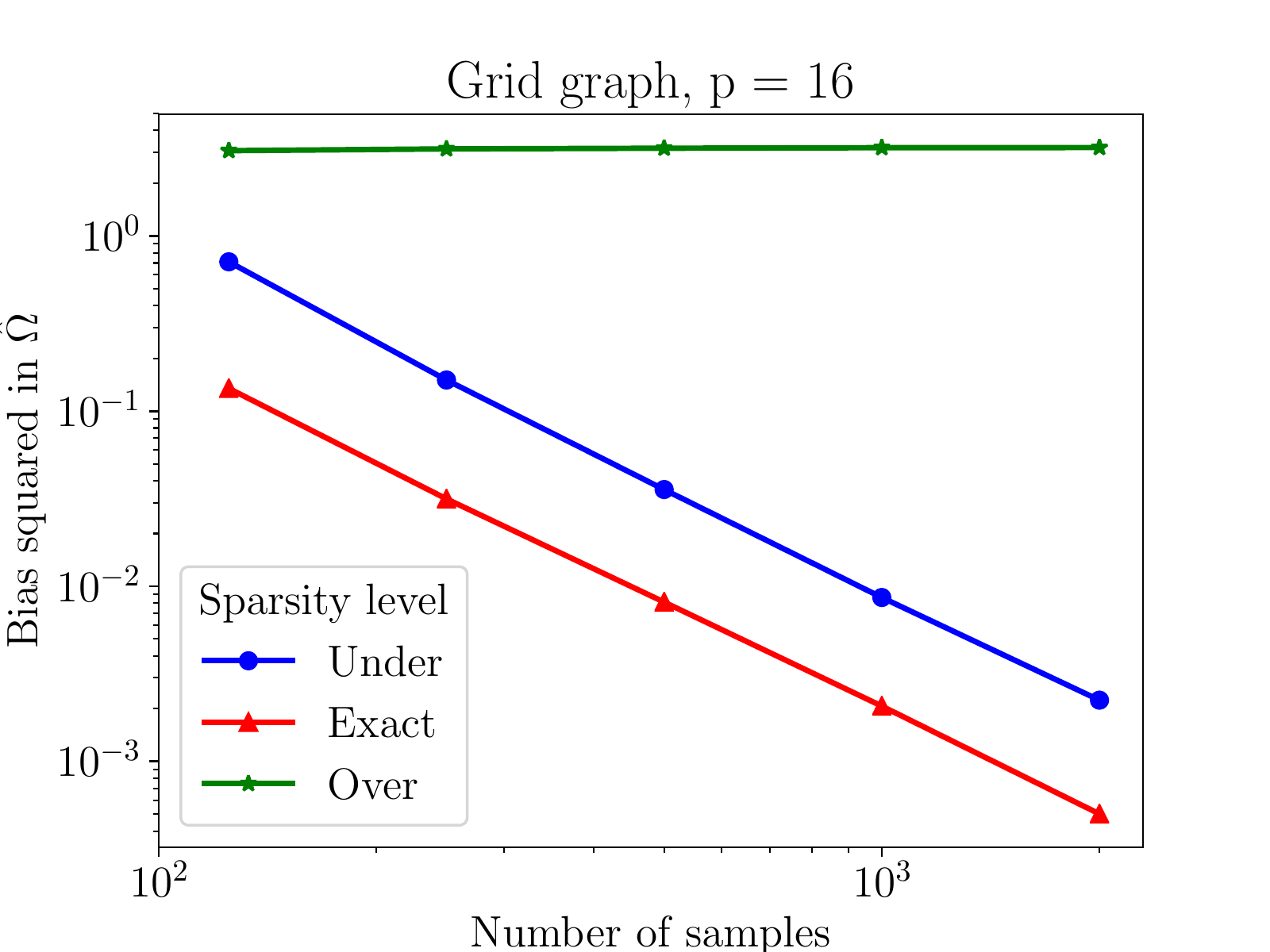}
        \caption{}
    \label{fig:bias} \end{subfigure}
    \caption{(a) Variance of $\hat{\Omega}_{jk}$ decreases with fewer coefficients and/or more samples; (b) Bias in
    $\hat{\Omega}_{jk}$ occurs with oversparsification. The bias and variance of $\hat{\Omega}$ are
computed using the Frobenius norm.}\label{fig:var}
\end{figure}

The critical point here is that sparsity in the graph implies sparsity in the map. The space of maps
that respect this sparsity pattern is denoted $\mathcal{S}_{\mathcal{I}}^\beta$. A sparser map
can be described by fewer coefficients $\bm{\alpha}$, which in turn decreases their total variance when
found via MLE. This improves the subsequent estimate of $\Omega$.
Numerical results supporting this claim are shown in Figure~\ref{fig:var} for a
Gaussian grid graph, $p=16$.
The plots show three levels of sparsity: ``under,'' corresponding to
a dense lower triangular map; ``exact,'' in which the map includes only the necessary active
variables; and ``over,'' corresponding to a diagonal map. In each case, the variance
decreases with increasing sample size, and the sparser the map, the lower the variance. However,
non-negligible bias is incurred when the map is over-sparsified; see Figure~\ref{fig:bias}. Ideally, the algorithm would move
from the under-sparsified level to the exact level.
\vspace{-.3cm}
\section{Algorithm: \textsc{SING}}\label{sec:sing}
We now present the full algorithm. Note that the ending condition is controlled by a variable
\texttt{DECREASING}, which is set to true until the size of the recovered edge set is no longer
decreasing.
The final ingredient is the thresholding step, explained in \S\ref{ssec:thresh}. Subscripts $l$ in
the algorithm refer to the given quantity at that iteration.

\IncMargin{1.2em}
    \begin{algorithm}[H]
    \caption{Sparsity Identification in Non-Gaussian distributions (\textsc{SING})}
\DontPrintSemicolon
        \SetKwInOut{Input}{input}\SetKwInOut{Output}{output}\SetKwInOut{Define}{define}
\BlankLine
        \Input{$n$ i.i.d. samples $\{\bm{x}^{(i)}\}_{i=1}^n \sim \pi$, maximum polynomial degree $\beta$} 
        \Output{ sparse edge set $\hat{E}$}
\BlankLine
        \Define{ $\mathcal{I}_{S_1} = \{\emptyset\}$, $l=1$,
        $|\hat{E}_0| = p(p-1)/2$, \texttt{DECREASING = TRUE}}
        \While{\texttt{DECREASING = TRUE}}{
            Estimate transport map $S_l \in \mathcal{S}_{\mathcal{I}_l}^\beta$, where $S_{l_\sharp} \pi = \eta$\;
        Compute $(\hat{\Omega}_l)_{jk} = \frac{1}{n}\sum_{i=1}^n\left|\partial_{jk} \log S_{\bm{\alpha}}^\sharp
                          \eta\left(\bm{x}^{(i)}\right)
                          \right|$\;
    Threshold $\hat{\Omega}_l$\;
        Compute $|\hat{E}_l|$ (the number of edges in the thresholded graph)\; 
  \If {$|\hat{E}_l| < |\hat{E}_{l-1}|$}{
      Find appropriate permutation of variables $\varphi_l$ (for example, reverse Cholesky ordering)\;
        Identify sparsity pattern of subsequent map $\mathcal{I}_{S_{l+1}}$\;
    $l \leftarrow l+1$\;
  }
  \Else{
      \texttt{DECREASING = FALSE}}
}
\end{algorithm}
\DecMargin{1.2em}
\textsc{SING} is not a strictly greedy algorithm---neither for the sparsity
pattern of the map nor for the edge removal of the graph.
First, the process of identifying the induced graph may involve
fill-in, and the extent of this fill-in might be larger than optimal due to the ordering heuristics.
Second, the estimate of the generalized precision is noisy due to finite sample size, and this noise
can add randomness to a thresholding decision. As a result, a variable that is set as inactive may
be reactivated in subsequent iterations.
%
However, we have found that oscillation in
the set of active variables is a rare occurence. Thus, checking that the total number of edges is
nondecreasing (as a global measure of sparsity) works well as a practical stopping criterion.
\subsection{Thresholding the generalized precision}\label{ssec:thresh}

An important component of this algorithm is a thresholding of the generalized precision. Based on literature
\citep{cai2011adaptive} and numerical results, we model the threshold as
$\tau_{jk} = \delta \rho_{jk}$, where $\delta$ is a tuning
parameter and $\rho_{jk} = [\mathbb{V}(\hat{\Omega}_{jk})]^{1/2}$ (where $\mathbb{V}$
denotes variance). Note that a threshold $\tau_{jk}$ is computed at each iteration and for every
off-diagonal entry of $\Omega$. More motivation for this choice is given in the scaling analysis of
the following section. The expression~(\ref{eq:mle-norm}) yields an estimate of the variances of the map
coefficients $\hat{\bm{\alpha}}$, but this uncertainty must still be propagated to the entries of
$\Omega$ in order to compute $\rho_{jk}$. This is possible using the delta method
\citep{oehlert1992note}, which states that if a sequence of one-dimensional
random variables satisfies
\[\sqrt{n}\, \left| X^{(n)} - \theta \right| \overset{d}{\longrightarrow}
\mathcal{N}\left(\mu, \sigma^2\right),\] then for a function $g(\theta)$, \[\sqrt{n}\, \left|
g\left(X^{(n)}\right) - g\left(\theta\right) \right| \overset{d}{\longrightarrow}
\mathcal{N}\left(g(\mu), \sigma^2|g'(\theta)|^2\right).\]
The MLE result also states that the coefficients are normally distributed as $n \rightarrow \infty$. Thus,
generalizing this method to vector-valued random variables gives an estimate for the variance in the entries of
$\Omega$, as a function of $\bm{\alpha}$, evaluated at the true minimizer $\bm{\alpha}^*$:
\begin{equation}\rho_{jk}^2 \approx \left(\nabla_{\bm{\alpha}} \Omega_{jk} \right)^T
    \left(\frac{1}{n} I^{-1}(\bm{\alpha})\right) \left(\nabla_{\bm{\alpha}}
\Omega_{jk}\right)\Big|_{\bm{\alpha}^*} .\label{eq:var-est}\end{equation}

\section{Scaling analysis}\label{sec:scale}

In this section, we derive an estimate for the number of samples needed to recover the exact graph
with some given probability. We consider a one-step version of the algorithm, or in other
words: what is the probability that the correct graph will be returned after a single step of
\textsc{SING}?
We also assume a particular instantiation of the transport map, and that $\kappa$, the
minimum non-zero edge weight in the true generalized precision, is given. That is, $\kappa =
\min_{j \neq k, \Omega_{jk} \neq 0}\left(\Omega_{jk}\right)$.

There are two possibilities for each pair $(j,k), j<k$: the edge $e_{jk}$ does exist in the true
edge set $E$ (case 1), or it does not (case 2). In case 1, the estimated value should be greater than its
variance, up to some level of confidence, reflected in the choice of $\delta$: $\Omega_{jk} >
\delta \rho_{jk}$. In the worst case, $\Omega_{jk} = \kappa$, so it must be that
$\kappa > \delta \rho_{jk}$. On the other hand, in case 2, in which the edge does not exist, then
similarly $\kappa - \delta \rho_{jk} > 0$.

If $\rho_{jk} < \kappa/\delta$, then by equation~(\ref{eq:var-est}), we have
\begin{equation}\frac{1}{n} \left(\nabla_{\bm{\alpha}} \Omega_{jk}\right)^T
    I^{-1}(\bm{\alpha})\left(\nabla_{\bm{\alpha}} \Omega_{jk}\right) <
\left(\frac{\kappa}{\delta}\right)^2\end{equation}
and so it must be that the number of samples\begin{equation} n  > \left(\nabla_{\bm{\alpha}}
    \Omega_{jk}\right)^T I^{-1}(\bm{\alpha})\left(\nabla_{\bm{\alpha}} \Omega_{jk}\right)
\left(\frac{\delta}{\kappa}\right)^2.
\end{equation}
Let us define the RHS above as $n_{jk}^*$ and set $n^* = \max_{j \neq k}\left(n_{jk}^*\right)$.

Recall that the estimate in line~(\ref{eq:ci}) contains the absolute value of a normally distributed
quantity, known as a folded normal distribution. In case 1, the mean is bounded away from zero, and
with small enough variance, the folded part of this distribution is negligible. In case 2, the mean
(before taking the absolute value) is zero, and so this estimate takes the form of a half-normal
distribution.

Let us now relate the level of confidence as reflected in $\delta$ to the probability $z$ that an
edge is correctly estimated. We define a function for the standard normal (in case 1) $\phi_1:
\mathbb{R}^+ \to (0,1)$ such that $\phi_1(\delta_1) = z_1$ and its inverse $\delta_1 = \phi_1^{-1}
(z_1)$, and similarly for the half-normal with $\phi_2$, $\delta_2$, and $z_2$. Consider the event
$B_{jk}$ as the event that edge $e_{jk}$ is estimated incorrectly: \[B_{jk} = \left\{ \left((e_{jk}
\in E) \cap (\hat{e}_{jk} \notin \hat{E})\right) \cup\left((e_{jk} \notin E) \cap (\hat{e}_{jk} \in
\hat{E})\right) \right\}.\]  In case 1,\[ \delta_1 \rho_{jk} < \kappa \implies P(B_{jk}) <
\frac{1}{2}(1-z_1)\] where the factor of $1/2$ appears because this event only occurs when the
estimate is below $\kappa$ (and not when the estimate is high). In case 2, we have \[ \delta_2
\rho_{jk} < \kappa \implies P(B_{jk}) < (1-z_2).\] To unify these two cases, let us define $z$
where $1 - z = (1-z_1)/2$, and set $z = z_2$.  Finally, we have $(B_{jk}) < (1-z),\,\,
j<k$.

Now we bound the probability that at least one edge is incorrect with a union bound:
\begin{align}
    P\left(\bigcup_{j<k} B_{jk}\right) &\leq \sum_{j<k}P(B_{jk})\\
     &= \frac{1}{2}p(p-1)(1-z).
\end{align}
Note $p(p-1)/2$ is the number of possible edges. The probability that an edge is incorrect increases
as $p$ increases, and decreases as $z$ approaches 1.  Next, we bound this probability of
recovering an incorrect graph by $m$. Then
$p(p-1)(1-z) < 2m$ which yields $z > 1 - 2m/\left(p(p-1)\right)$. Let
\begin{align}
    \delta^* = \max\left[\delta_1, \delta_2\right] &= 
    \max\left[\phi_1^{-1}\left(1 - \frac{2m}{p(p-1)}\right), \phi_2^{-1}\left(1 -
    \frac{2m}{p(p-1)}\right)\right].
\end{align}
Therefore, to recover the correct graph with probability $m$ we need at least $n^*$ samples, where
\[n^* = \max_{j \neq k} \left\{ \left(\nabla_{\bm{\alpha}} \Omega_{jk}\right)^T
I^{-1}(\bm{\alpha})\left(\nabla_{\bm{\alpha}} \Omega_{jk}\right)
\left(\frac{\delta^*}{\kappa}\right)^2\right\}.\]

\section{Examples}\label{sec:ex}


\subsection{Modified Rademacher}\label{ssec:mr}
\begin{figure}[!htb]
  \centering
  \begin{subfigure}{.2\textwidth}
    \centering
    \includegraphics[width=\textwidth]{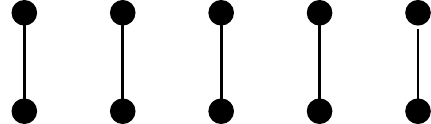}
      \caption{}
  \end{subfigure}
  \begin{subfigure}{.4\textwidth}
  \centering
    \includegraphics[width=\textwidth]{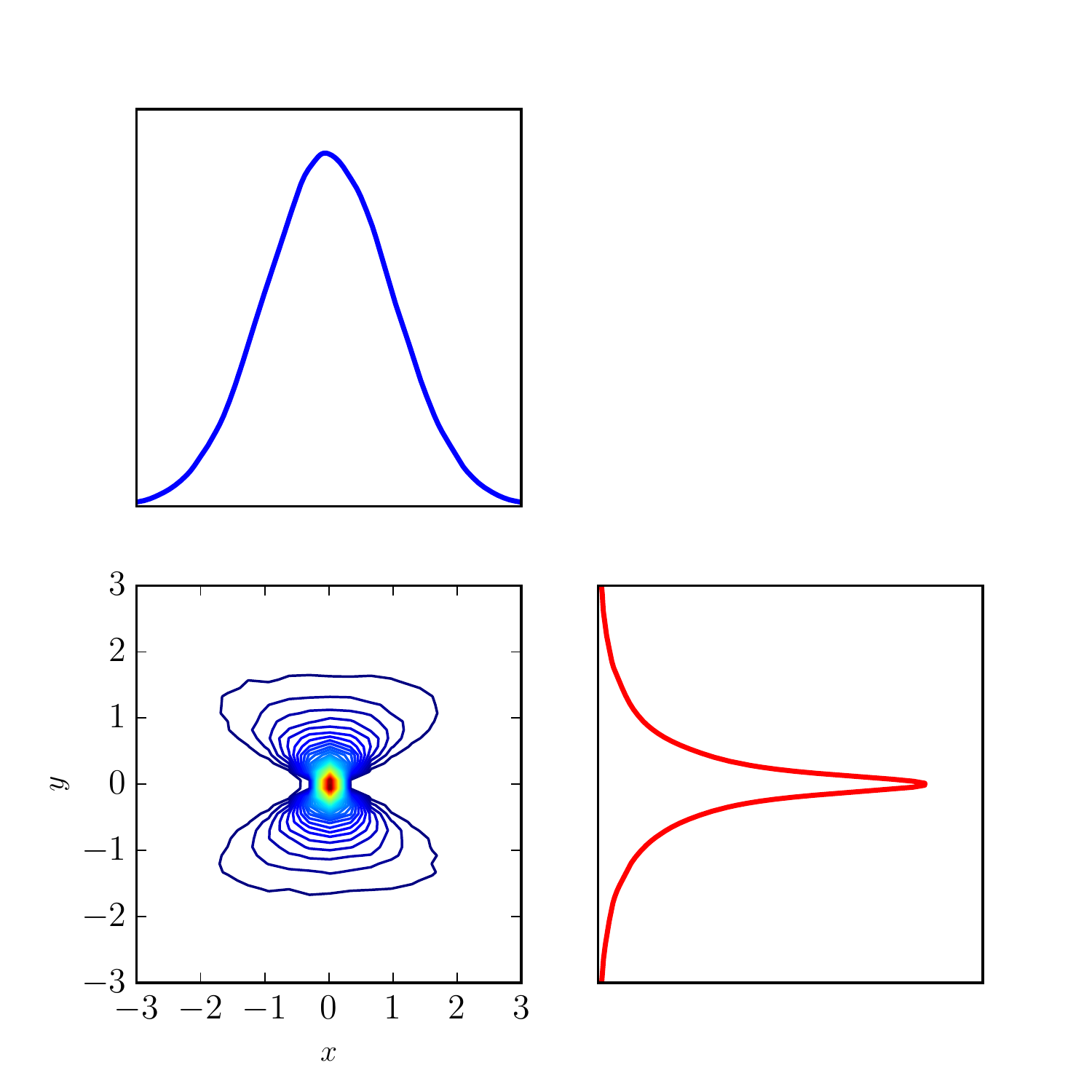}
      \caption{}
  \end{subfigure}\hspace{-1em}
  \begin{subfigure}{.33\textwidth}
   \centering
   \includegraphics[width=\textwidth]{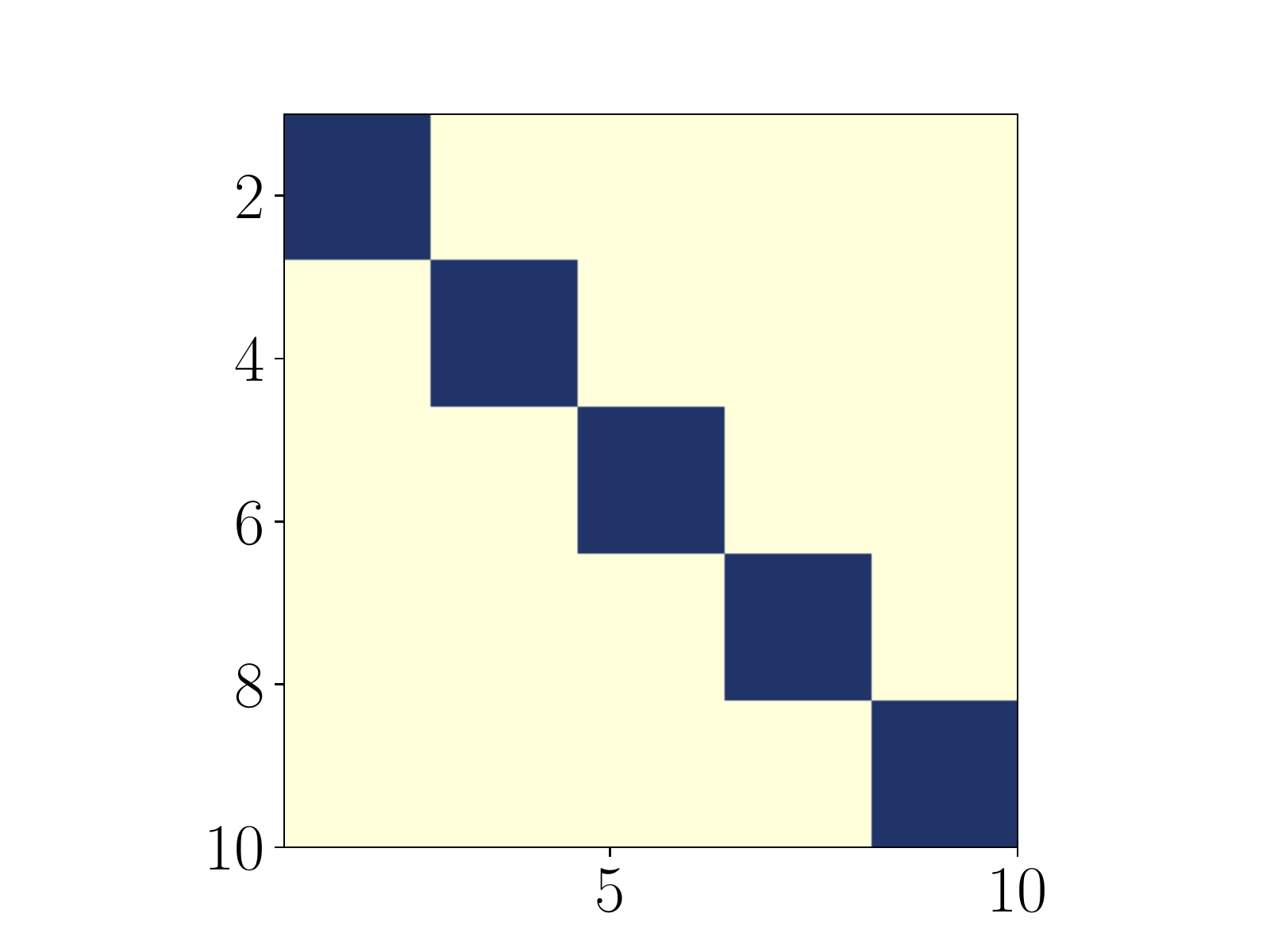}
      \caption{}
  \end{subfigure}\hspace{-1em}
\caption{(a) The undirected graphical model; (b) One- and two-dimensional marginal distributions
    for one pair $(X,Y)$; (c) Adjacency matrix of true graph (dark blue corresponds to
    an edge, off-white to no edge).\label{fig:modrad}}
\end{figure}

Consider $r$ pairs of random variables $(X,Y)$, where:
\begin{align}
    X &\sim \mathcal{N}(0,1)\\
    Y &= WX, \quad \text{with } W \sim \mathcal{N}(0,1).
    \end{align} 
(A common example illustrating that two random variables can be uncorrelated but not independent uses
draws for $W$ from a Rademacher distribution, which are $-1$ and $1$ with equal probability.)
When $r=5$, the corresponding graphical model and support of the generalized precision are shown in
Figure~\ref{fig:modrad}. The same figure also shows the one- and two-dimensional marginal
distributions for one pair $(X,Y)$. Each 1-dimensional marginal is symmetric and unimodal, but the
two-dimensional marginal is quite non-Gaussian.

\begin{figure}[!htb]
  \centering
  \begin{subfigure}{.22\textwidth}
   \centering
      \includegraphics[width=\textwidth,trim={2cm 0 2 0},clip]{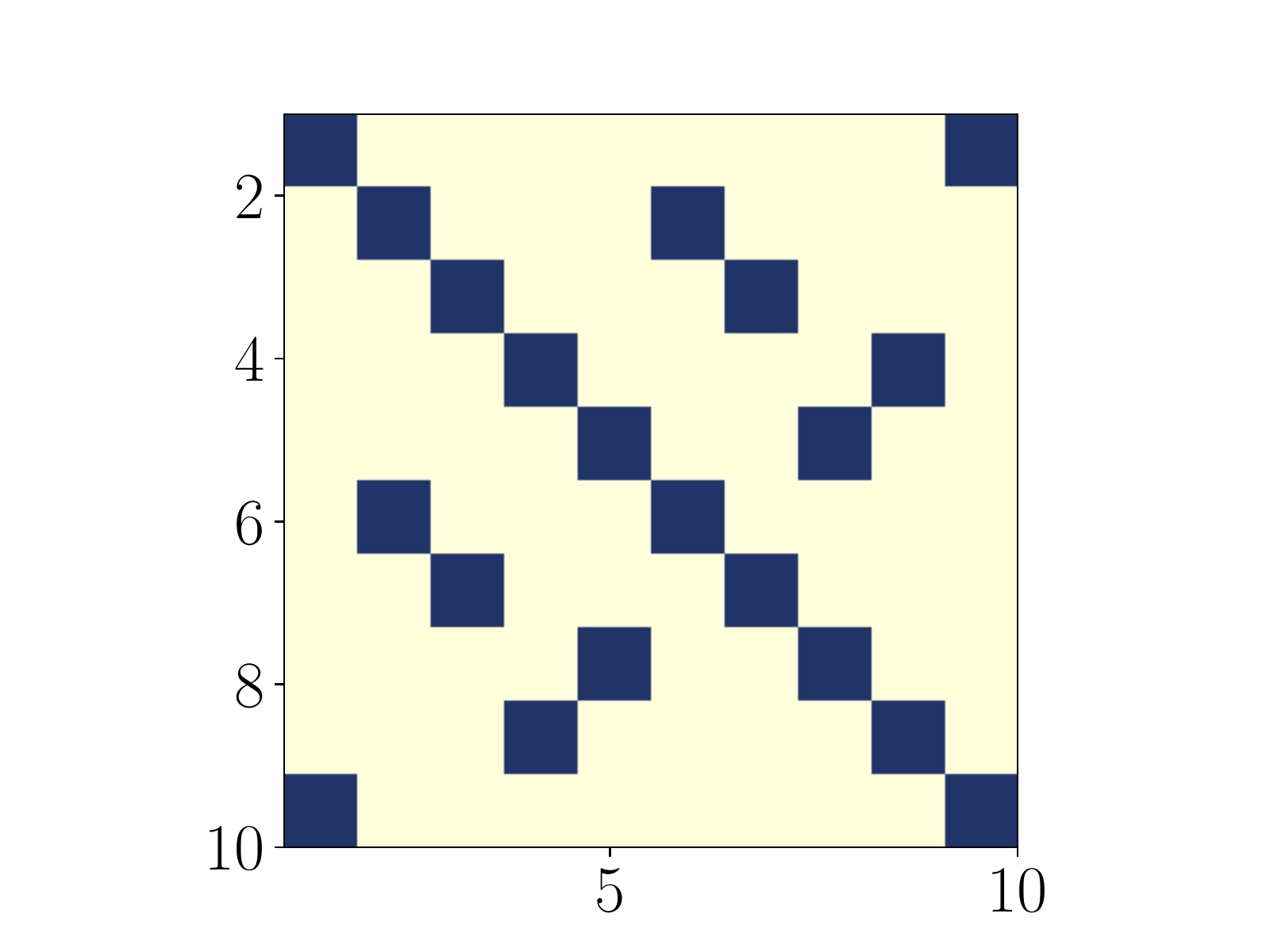}
      \caption{\label{fig:modrad-a}}
  \end{subfigure}\hspace{-1em}
  \begin{subfigure}{.22\textwidth}
   \centering
   \includegraphics[width=\textwidth,trim={2cm 0 2 0},clip]{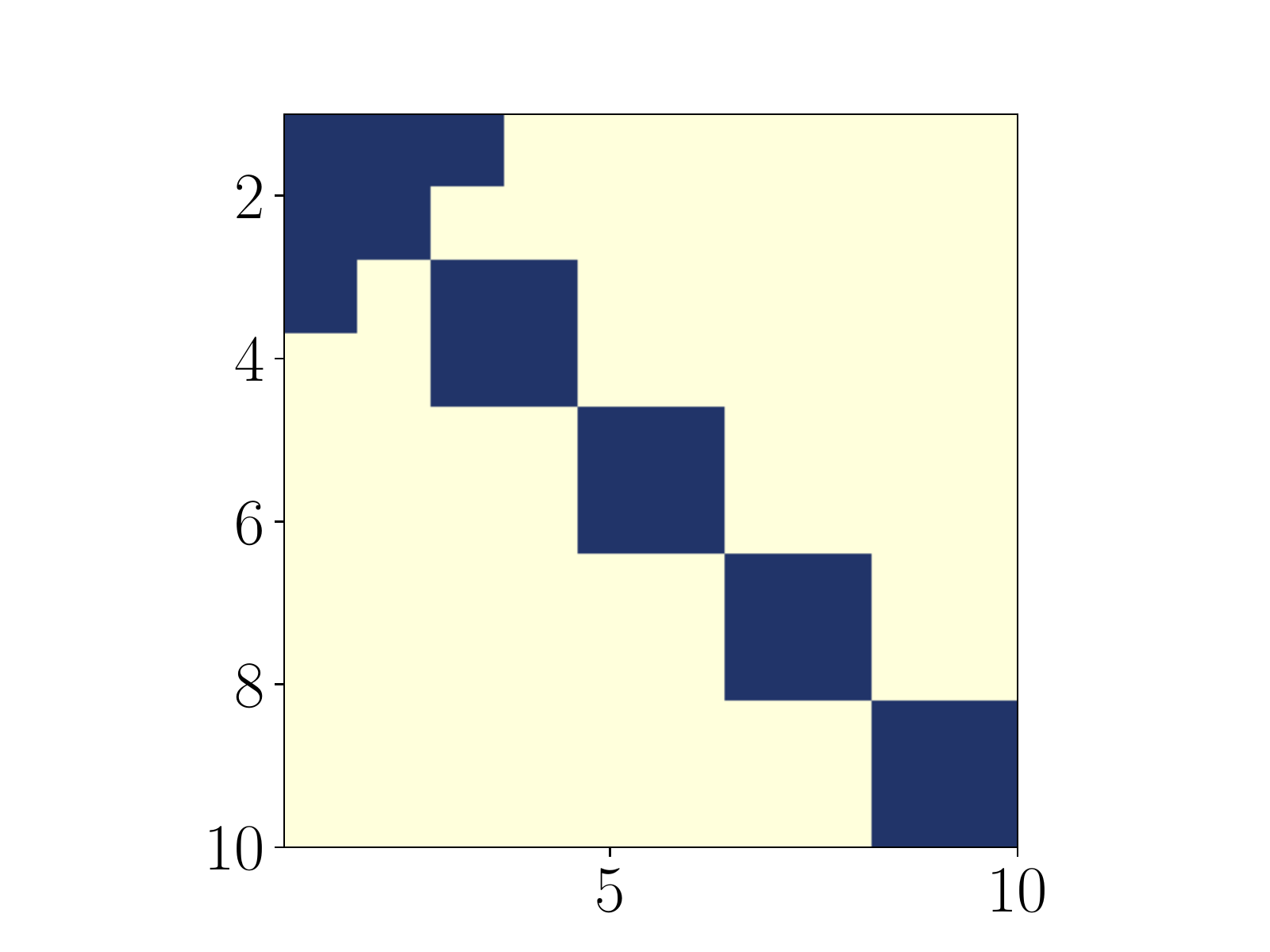}
      \caption{\label{fig:modrad-b}}
  \end{subfigure}\hspace{-1em}
  \begin{subfigure}{.22\textwidth}
  \centering
   \includegraphics[width=\textwidth,trim={2cm 0 2 0},clip]{matrix-slots-mr-3-nips.pdf}
      \caption{\label{fig:modrad-c}}
\end{subfigure}\hspace{-1em}
  \begin{subfigure}{.22\textwidth}
  \centering
  \includegraphics[width=\textwidth,trim={2cm 0 2 0},clip]{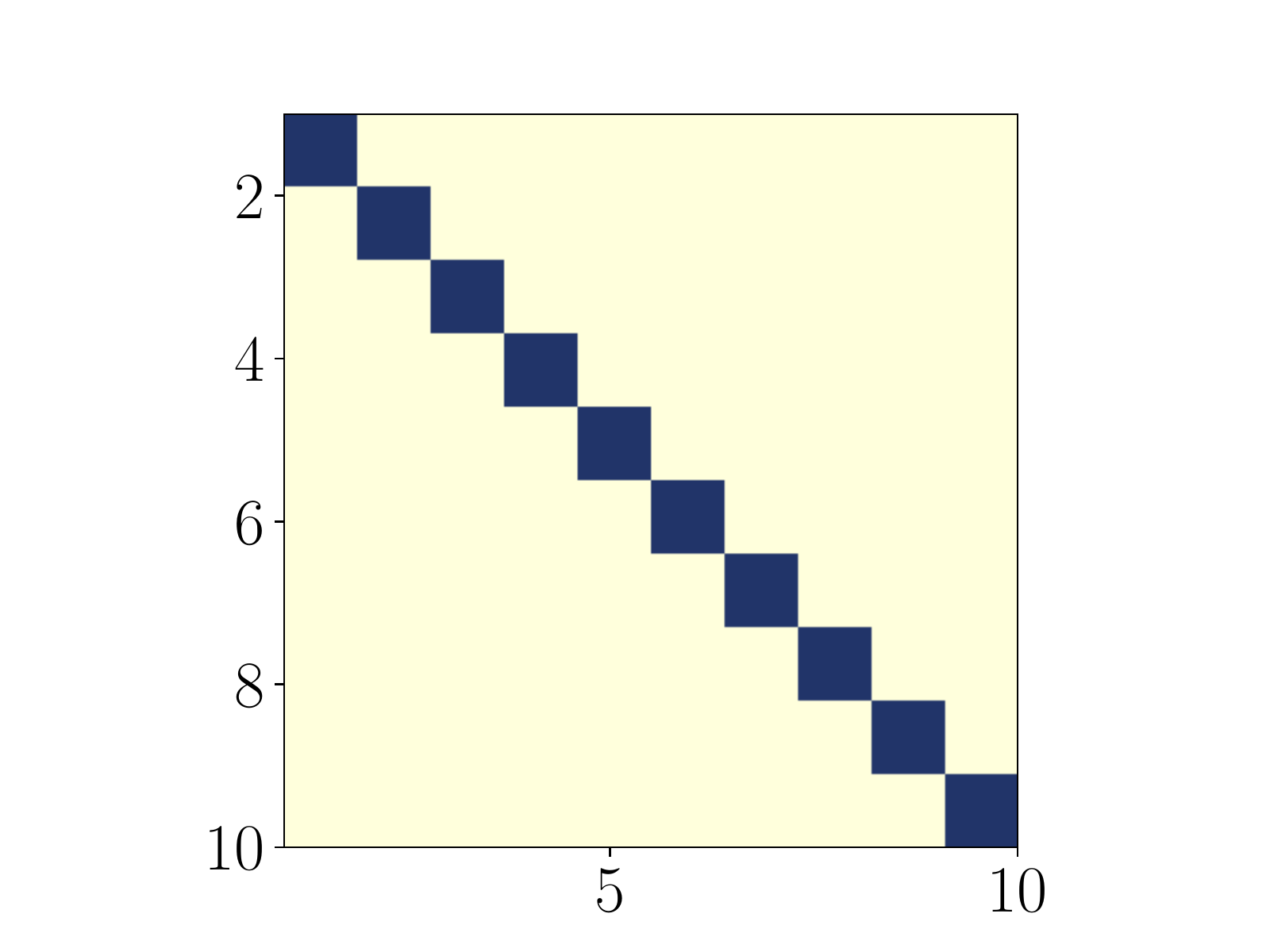}
    \caption{\label{fig:modrad-d}}
\end{subfigure}
    \caption{(a) Adjacency matrix of original graph under random variable permutation; (b) Iteration
    1; (c) Iterations 2 and 3 are identical: correct graph recovered via \textsc{SING} with $\beta=2$; (d) Recovered graph, using \textsc{SING} with $\beta=1$.\label{fig:alg-mr}}
\end{figure}

Figures~\ref{fig:modrad-a}--\ref{fig:modrad-c} show the progression of the identified graph over the iterations of the
algorithm, with $n=2000$, $\delta=2$, and maximum degree $\beta=2$. The variables are initially permuted
to demonstrate that the algorithm is able to find a good ordering. After the first iteration, one extra
edge remains. After the second, the erroneous edge is removed and the graph is correct. After the
third, the sparsity of the graph has not changed and the recovered graph is returned as is.
Importantly, an assumption of normality on the data returns the incorrect graph, displayed in
Figure~\ref{fig:modrad-d}. (This assumption can be enforced by using a linear transport map, or
$\beta=1$.) In fact, not only is the graph incorrect, the use of a linear map fails to detect any
edges at all and deems the ten variables to be independent.

\subsection{Stochastic volatility}\label{ssec:sv}

\begin{figure}[h]
  \centering
    \begin{subfigure}{.33\textwidth}
  \includegraphics[width=\textwidth]{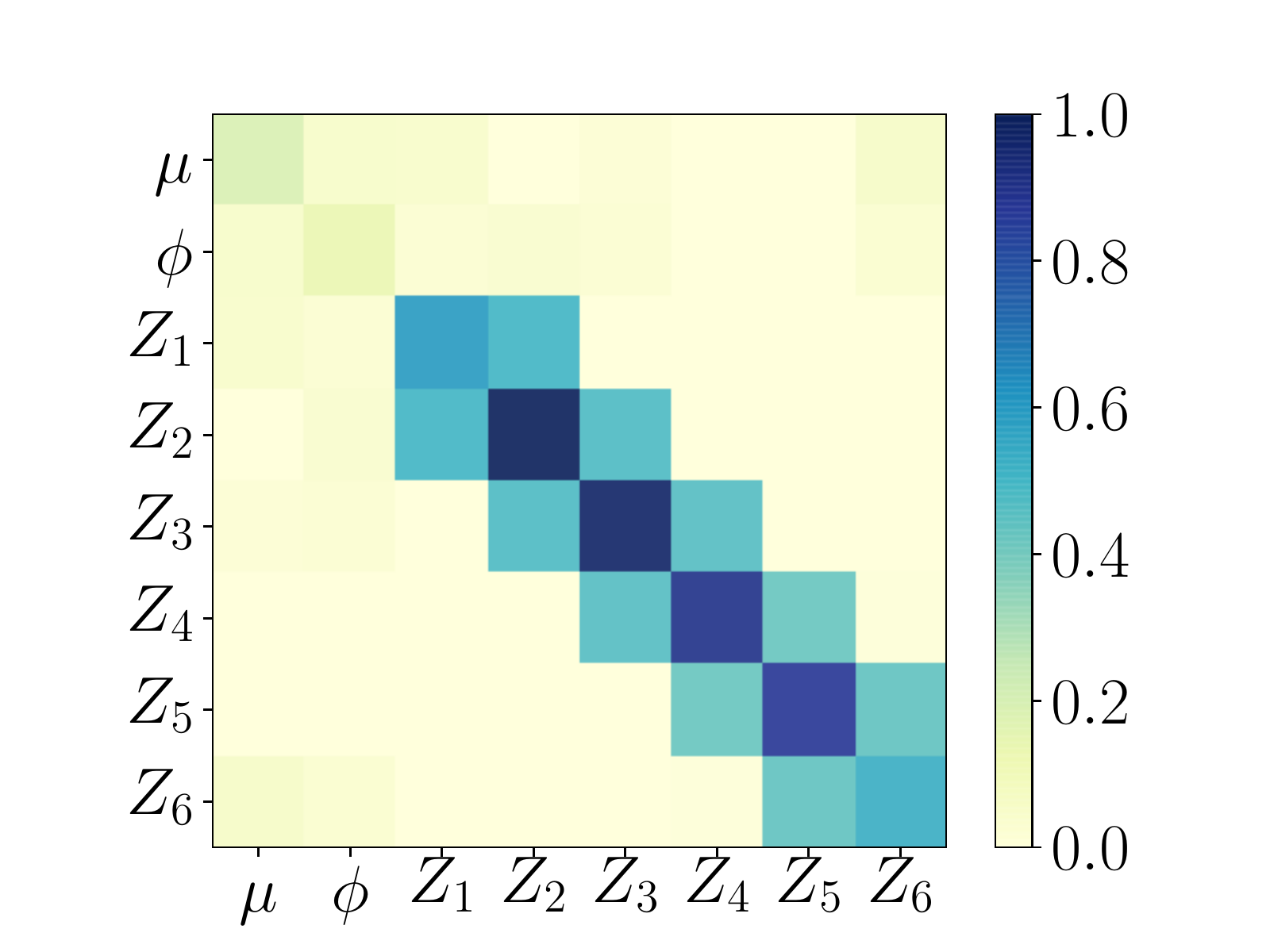}
        \caption{\label{fig:sv-gp}}
  \end{subfigure}\hspace{-1em}
    \begin{subfigure}{.33\textwidth}
  \includegraphics[width=\textwidth]{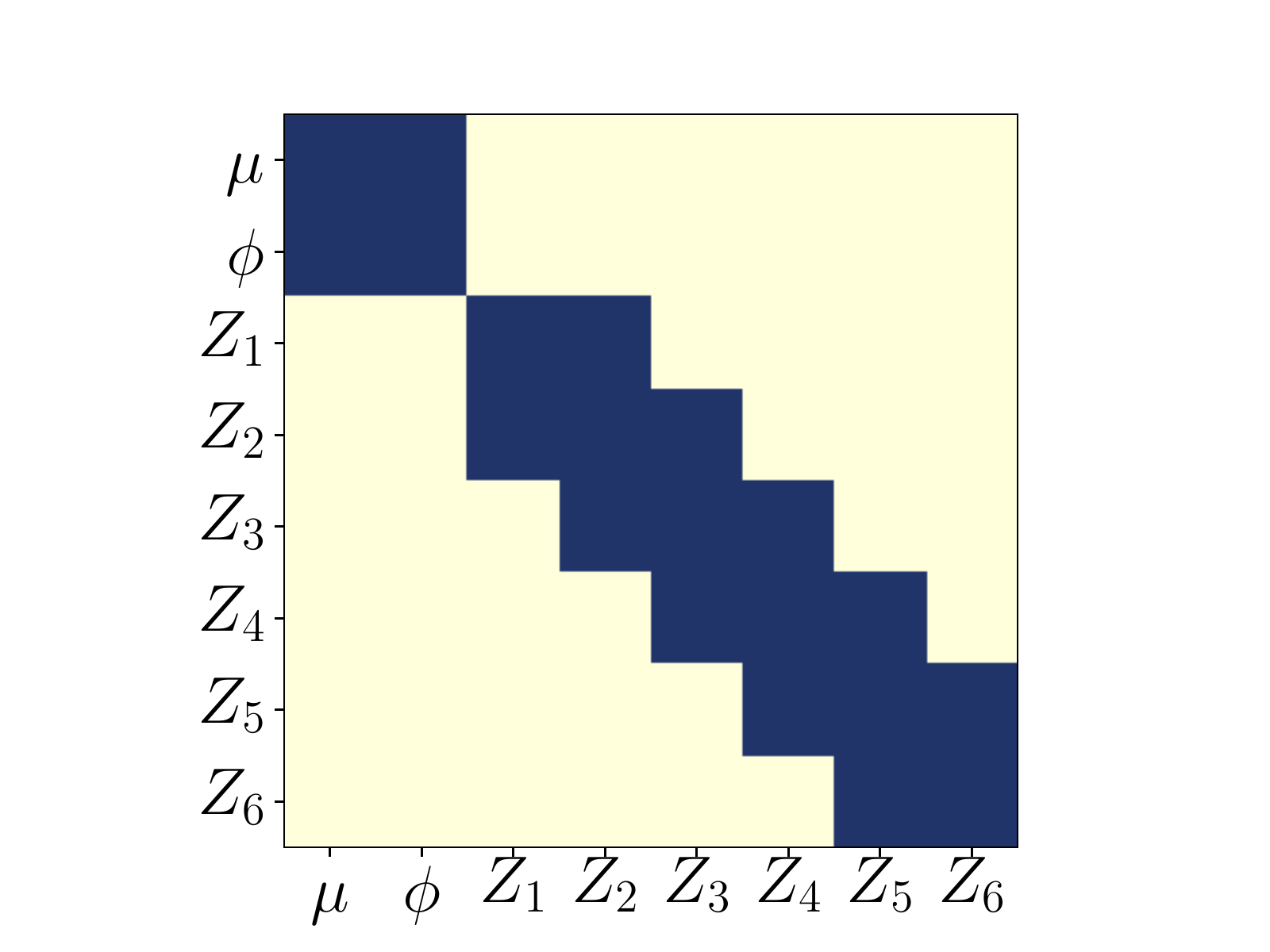}
        \caption{\label{fig:sv-strong}}
  \end{subfigure}\hspace{-1em}
    \caption{(a) The scaled generalized precision matrix
    $\hat{\Omega}$; (b)~Strong edges recovered via \textsc{SING}, $n=2000$.
    \label{fig:sv-slots}}
\end{figure}

As a second example, we consider data generated from a stochastic volatility model of a financial
asset \citep{rue2005gaussian, kim1998stochastic}. The log-volatility of the asset is modeled as an
autoregressive process at times $t = 1,\dots,T$. In particular, the state at time $t+1$ is given as
\begin{equation}
    Z_{t+1} = \mu + \phi(Z_t - \mu) + \epsilon_t, \quad \epsilon_t \sim \mathcal{N}(0,1)
\end{equation}
where
\begin{align}
    Z_0 | \mu, \phi &\sim \mathcal{N}\left( \mu, \frac{1}{1 - \phi^2} \right), \quad \mu
    \sim \mathcal{N}(0,1)\\
    \phi &= 2 \frac{e^{\phi^*}}{1 + e^{\phi^*}} - 1, \quad \phi^* \sim \mathcal{N}(3,1).
\end{align}
\begin{figure}
  \centering
    \begin{subfigure}{.45\textwidth}\hspace{-.8cm}
        \centering
  \includegraphics[width=\textwidth]{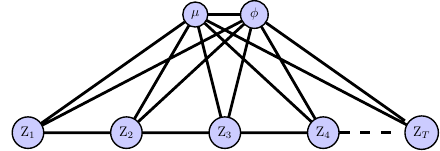}
        \caption{}
  \end{subfigure}\hspace{-1em}
    \begin{subfigure}{.33\textwidth}
        \centering
  \includegraphics[width=\textwidth]{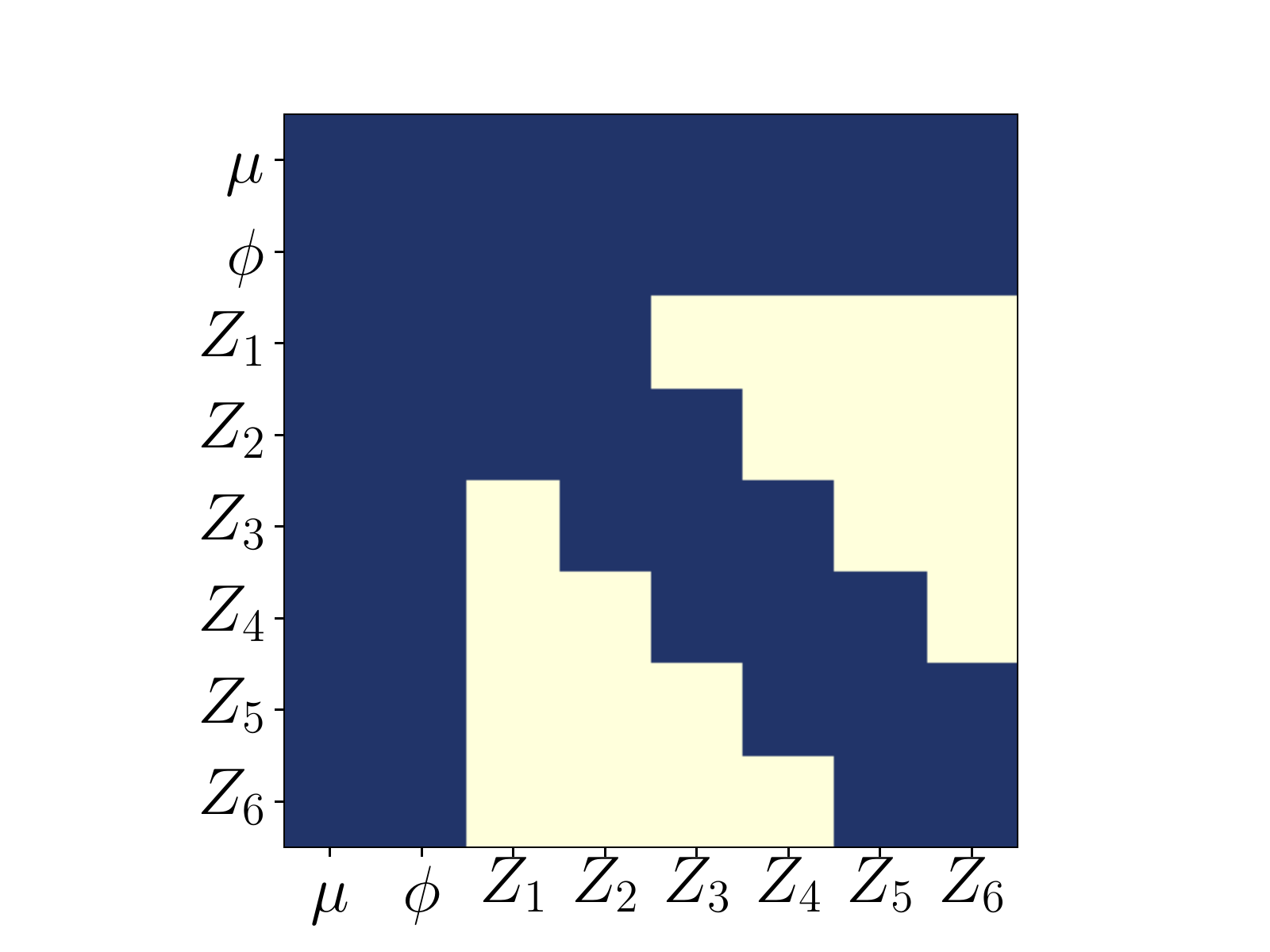}
        \caption{}
  \end{subfigure}\hspace{-1em}
    \caption{(a) The graph of the stochastic volatility model; (b) Adjacency matrix of true graph.
    \label{fig:sv}\vspace{-.3cm}}
\end{figure}%
\begin{figure}
  \centering
    \begin{subfigure}{.33\textwidth}
  \includegraphics[width=\textwidth]{matrix-slots-adj-sv-nips.pdf}
        \caption{\label{fig:sv-all}}
  \end{subfigure}\hspace{-1em}
    \begin{subfigure}{.33\textwidth}
        \centering
  \includegraphics[width=\textwidth]{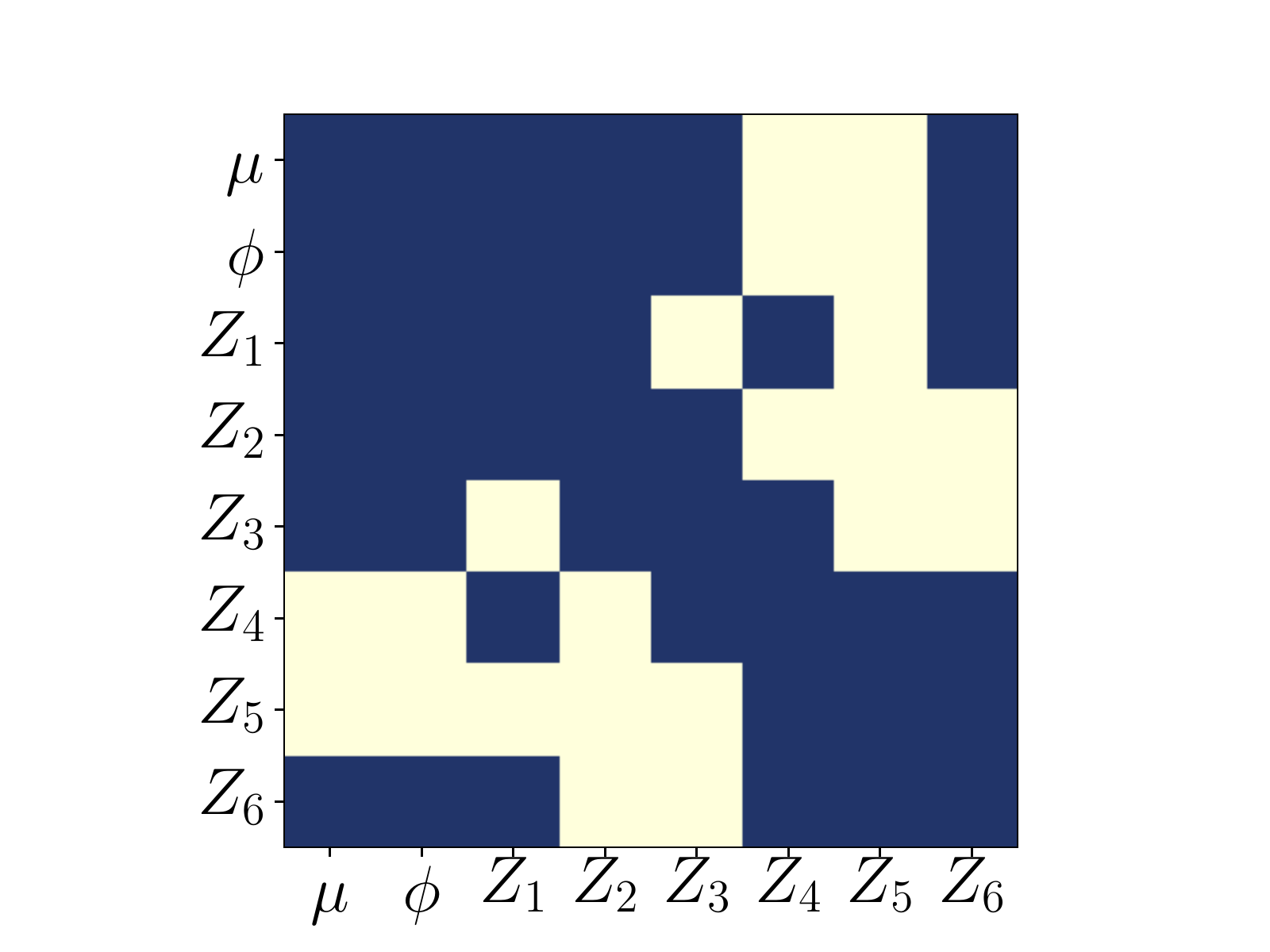}
        \caption{}
  \end{subfigure}\hspace{-1em}
    \begin{subfigure}{.33\textwidth}
        \includegraphics[width=\textwidth]{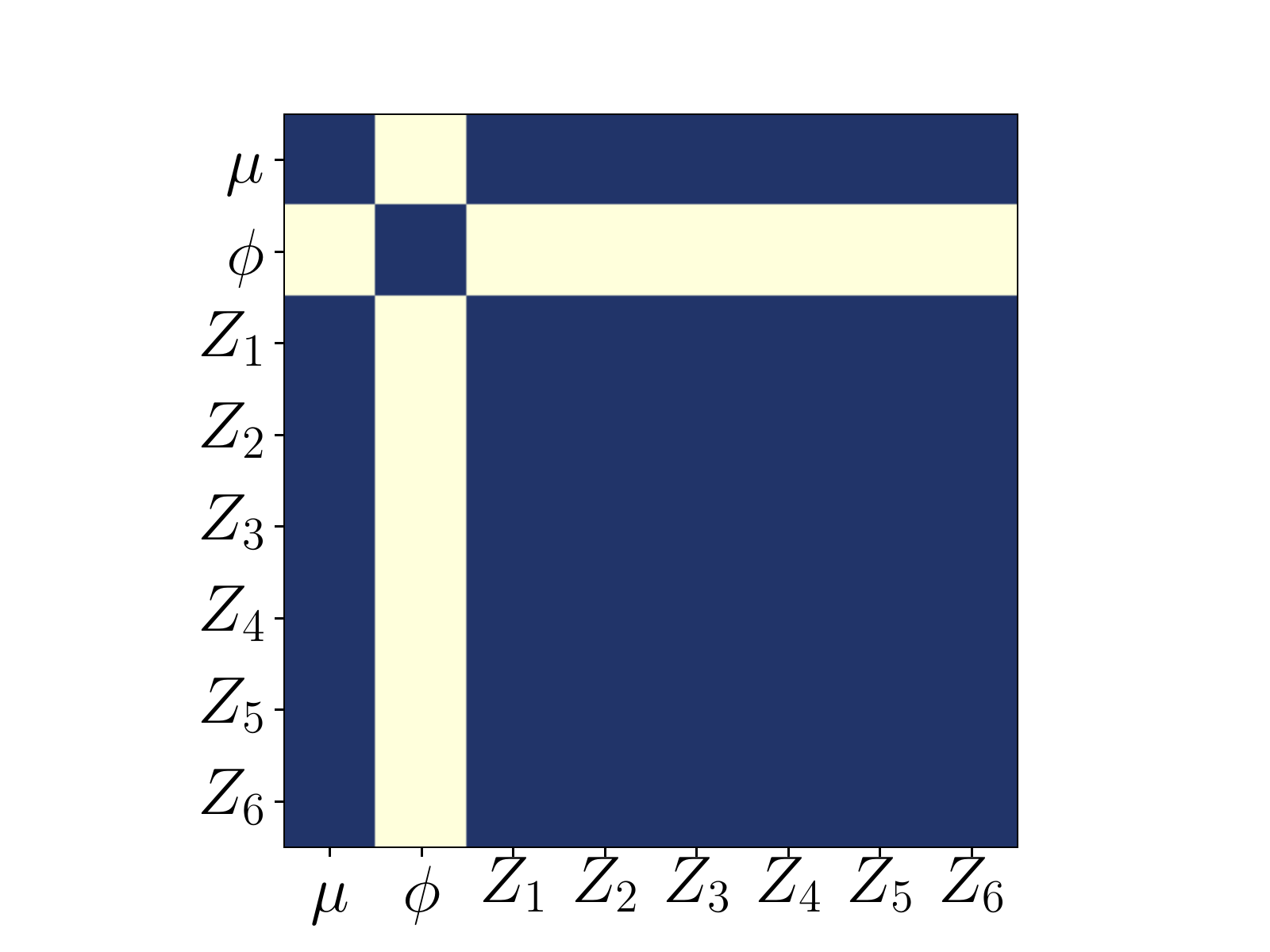}
        \caption{\label{fig:sv-glasso}}
    \end{subfigure}
    \caption{Recovered graphs using: (a)~\textsc{SING}, $\beta=2$, $n=15000$; (b)~\textsc{SING},
    $\beta = 1$; (c)~\textsc{GLASSO}.\label{fig:sv}\vspace{-.3cm}}
\end{figure}%
The corresponding graph is depicted in Figure~\ref{fig:sv}. With $T=6$, samples were generated from the
posterior distribution of the state $\bm{Z}_{1:6}$ and hyperparameters $\mu$ and
$\phi$, given noisy measurements of the state. Using a relatively large number of samples $n=15000$,
$\delta = 1.5$, and $\beta=2$, the correct graph is recovered, shown in Figure~\ref{fig:sv-all}.
With the same amount of data, a linear map returns the incorrect graph---having both missing and
spurious additional edges.  The large number of samples is required because the edges between hyperparameters
and state variables are quite weak.  Magnitudes of the entries of the generalized precision (scaled
to have maximum value $1$) are displayed in Figure~\ref{fig:sv-gp}. The stronger edges may be
recovered with a much smaller number of samples ($n=2000$), however; see Figure~\ref{fig:sv-strong}.
This example illustrates the interplay between the minimum edge weight $\kappa$ and the number of
samples needed, as seen in the previous section. In some cases, it may be more reasonable to expect
that, given a fixed number of samples, \textsc{SING} could recover edges with edge weight above some
$\kappa_{\text{min}}$, but would not reliably discover edges below that cutoff. Strong edges could
also be discovered using fewer samples and a modified \textsc{SING} algorithm with $\ell_1$
penalties (a modification to the algorithm currently under development).

For comparison, Figure~\ref{fig:sv-glasso} shows the graph produced by assuming that the data are Gaussian
and using the \textsc{GLASSO} algorithm \citep{friedman2008sparse}. Results were generated for 40 different
values of the tuning parameter $\lambda \in (10^{-6},1)$. The result shown here was chosen such that
the sparsity level is locally constant with respect to $\lambda$, specifically at $\lambda = .15$.
Here we see that using a Gaussian assumption with non-Gaussian data overestimates edges among state
variables and underestimates edges between state variables and the hyperparameter $\phi$.

\section{Discussion}

The scaling analysis presented here depends on a particular representation of the transport map.  An
interesting open question is: What is the information-theoretic (representation-independent) lower
bound on the number of samples needed to identify edges in the non-Gaussian setting? This question
relates to the notion of an {\it information gap}: any undirected graph satisfies the Markov
properties of an infinite number of distributions, and thus identification of the graph should
require less information than that of the distribution. Formalizing these notions is an important
topic of future work.

\subsubsection*{Acknowledgments}


This work has been supported in part by the AFOSR MURI on
  ``Managing multiple information sources of multi-physics systems,''
  program officer Jean-Luc Cambier, award  FA9550-15-1-0038.  We
  would also like to thank Daniele Bigoni for generous help with code
  implementation and execution.
\small

\bibliographystyle{abbrv}  




\end{document}